\RequirePackage{amsmath}
\documentclass[runningheads]{llncs}

\usepackage{amssymb}
\usepackage{amsmath}
\usepackage{graphicx}
\usepackage{url}
\usepackage{color}
\usepackage{subfig}
\usepackage{times}
\usepackage{paralist}

\newcommand{\citep}[1]{\cite{#1}}

\definecolor{red}{rgb}{0.894, 0.102, 0.11}
\definecolor{blue}{rgb}{0.216, 0.494, 0.722}
\definecolor{green}{rgb}{0.302, 0.686, 0.29}
\definecolor{purple}{rgb}{ 0.596, 0.306, 0.639}
\definecolor{orange}{rgb}{ 1.00, 0.498, 0}
\definecolor{gray}{rgb}{0.60, 0.60, 0.60}
\definecolor{brown}{rgb}{0.651, 0.337, 0.157}
\definecolor{pink}{rgb}{0.969, 0.506, 0.749}

\newcommand{\todo}[1]{{\leavevmode\color{red}#1}}
\renewcommand{\todo}[1]{}
\newcommand{\com}[1]{{\leavevmode\color{blue}#1}}
\renewcommand{\com}[1]{}
\newcommand{\raus}[1]{{\leavevmode\color{gray}#1}}
\renewcommand{\raus}[1]{}

\begin{document}

\title{Simplifying Random Forests: \\On the Trade-off between Interpretability and Accuracy}
\titlerunning{Simplifying Random Forests}

\author{Michael Rapp \and Eneldo Loza Menc\'ia \and Johannes F\"urnkranz}
\authorrunning{M. Rapp \and E. Loza Menc\'ia \and J. F\"urnkranz}

\institute{Knowledge Engineering Group, TU Darmstadt, Germany}

\maketitle

\begin{abstract}
We analyze the trade-off between model complexity and accuracy for random forests by breaking the trees up into individual classification rules and selecting a subset of them. We show experimentally that already a few rules are sufficient to achieve an acceptable accuracy close to that of the original model. Moreover, our results indicate that 
in many cases, this can lead to simpler models that clearly outperform the original ones.
\end{abstract}

\section{Introduction}

Random forests (RF) \citep{random_forest} are a state-of-the-art method for solving classification and regression tasks in a wide range of domains \citep{delgado14classifiercomparison}. This is even the case for tasks for which it is commonly accepted that decision trees are inappropriate models, e.g., in domains with continuous input spaces for which the axis-aligned decision boundaries of regular decision trees impose a limitation. Two aspects in the induction of RF seem to be crucial for that improved ability. First, due to random feature sub-sampling and bagging, which leads to slightly shifted class distributions, each decision tree selects slightly different feature thresholds, patterns and conditions. 
\raus{
Firstly, the induction of each decision tree in a RF is perturbed in a way that makes it necessary to find alternative feature patterns (due to random feature sub-sampling) and alternative conditions on the features (due to slightly shifted class distributions or feature thresholds due to bagging).
}
This results in a variety of alternative explanations, increasing the chances of including patterns with a good generalization and simultaneously avoiding overfitting by maintaining a certain degree of variability. 
Second, by using an ensemble of trees whose predictions are combined via voting, random forests achieve smooth decision boundaries which enables them to address problems not solvable by 
common decision tree learner for single trees
\citep{wyner17explainingRF}.

Unfortunately, one of the key advantages of decision trees, namely their interpretability, is strongly limited. Even though each of the trees can be comprehended and analyzed individually, inspecting hundreds or thousands of them would easily overwhelm a human being. 
In this work, we are interested in the trade-off between the effectiveness and the interpretability of such ensembles. 
We propose to transform random forests into classification rules and investigate said trade-off by selecting subsets of these rules using different approaches and heuristics known from the field of rule induction. In our analysis we consider different aspects such as coverage and completeness, redundancy avoidance, and the accuracy of the final models.
We argue that a rule-based strategy has conceptual advantages compared to the selection of a subset of trees. 
First, it allows a more fine-grained analysis since we can investigate the effect of adding individual rules rather than entire trees that potentially consists of a large number of rules. Second, to base the selection on trees rather than rules would not result in the best trade-off between effectiveness and interpretability, as each tree covers the full instance space and therefore contributes a large number of redundant rules to the model.


\raus{
We analyze different selection heuristics and approaches from the field of rule induction with respect to different aspects such as coverage and completeness, redundancy avoidance and accuracy of the final model. 
}
\raus{
The experimental evaluation shows that the right selection process is crucial when breaking down forests into rules.
Moreover, we not only show that only a few rules are sufficient in order to obtain satisfactory accuracy, but also that the right selection of the rules can greatly improve over using all rules.\com{the last two sentences can be left out since we already say this in the abstract and will say it later on in the conclusions (for sb quickly just reading abstract, intro, conclusions).}
}
\section{Simplifying Random Forests by Selection of Rules}

\raus{
\section{Random Forests and Extraction of Rules}

A random forest consists of a predefined number of decision trees. For building the individual trees, only a subset of the available training instances and attributes are taken into account, which guarantees a diverse set of trees. Bagging is used for sampling the instances, i.e., a specific number of instances is drawn randomly with replacement. Additionally, each times a new node is added to one of the trees, only a random selection of the available features is considered.

A decision tree can be converted into an equivalent set of rules $\boldsymbol{r}: H \leftarrow B$ by considering each path from the root node to a leaf as a propositional rule. The body $B$ of such a rule consists of a conjunction of all conditions encountered on the respective path, whereas the head $H$ predicts the class associated with the corresponding leaf.

When using a RF to make a prediction, each of its decision trees provides a prediction for the given instance. To obtain a final result, these predictions are usually aggregated via voting. When converting the decision trees of a RF into rules, this is equivalent to a vote among all rules that cover the given instance, i.e., the rules for which all conditions in the body are satisfied.

\section{Selection of Rules}
}
A random forest consists of a predefined number of $i$ decision trees. For building the individual trees, only a subset of the available training instances (bagging) and attributes (random feature sampling) are considered, which guarantees a diverse set of trees \citep{random_forest}.

The trees of a RF can be converted into an equivalent set of $d$ rules $\boldsymbol{r}: H \leftarrow B$ by considering each path from the root node to a leaf as a propositional rule, with the conjunction $B$ containing all conditions encountered on the path and $H$ predicting the class in the leaf. 
For making a prediction, each covering rule provides a vote for the class in its head. Note that the way these rules are generated guarantees that each example will be covered by exactly $i$ rules.

To assess the quality of individual rules, we rely on heuristics commonly used in rule learning (see e.g., \citep{furnkranz2012}).  These heuristics map the number of true positives, false positives, true negatives, and false negatives predicted by a rule into a heuristic value $h$, typically $\in \left[ 0, 1 \right]$. Among the heuristics we use are \emph{precision} or \emph{confidence}, which 
has frequently been used
despite its tendency to overfit, as well as \emph{recall}, which we expect to work well in a setting where voting is used. Additionally, we use a parametrization of the \emph{m-estimate}, which has proved to be one of the most effective heuristics in rule learning \citep{janssen2010}.
%
%
For selecting a subset of $n < d$ rules among all $d$ rules that have been extracted from a RF, we use the following strategies.
\raus{
\begin{itemize}
    \item \textbf{Best rules:} 
    All rules are evaluated in terms of a certain heuristic and sorted by their heuristic values in decreasing order. The first $n$ rules are included in the subset.
    \item \textbf{Weighted covering:} An iterative approach, where rules are re-validated with respect to previously selected rules. In each step, the rule with the greatest heuristic value is selected and the weights of the instances it covers are halved. By recalculating the heuristic values of the remain rules based on the new weights, equal coverage of the training instances is achieved. The process continues until $n$ rules have been selected.
\end{itemize}
}
\begin{inparaenum}[\bfseries(1)]
  \item \textbf{Best rules:} The best $n$ rules according to a specified heuristic are included in the subset.
  \item \textbf{Weighted covering:} Like (1), but after selecting a rule, 
 the weights of the instances it covers are halved. By revalidating the remaining rules based on the new weights equal coverage of the training data is achieved. 
\end{inparaenum}

\section{Experiments}

For our experiments, we trained a random forest with 100 trees on the training data set, extracted $n$ rules with respect to the strategies mentioned above, and evaluated their accuracy on the test set.

\begin{figure}[t]
    \centering
\subfloat[30 best rules \label{fig:synth30best}]{\includegraphics[width=0.32\linewidth,height=32mm]{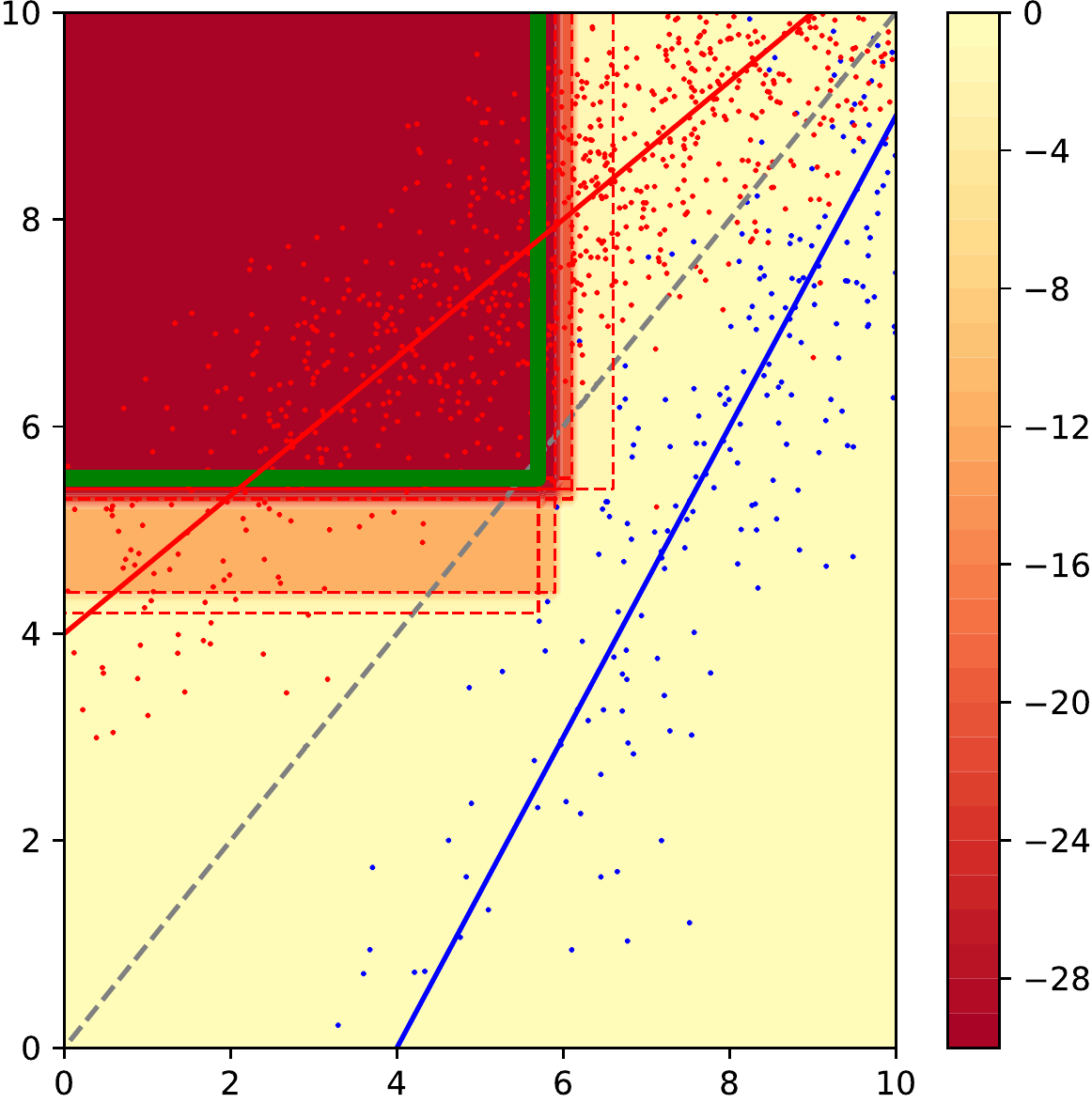}}\hfill%
\subfloat[30 weighted covering\label{fig:synth30wcov}]{\includegraphics[width=0.32\linewidth,height=32mm]{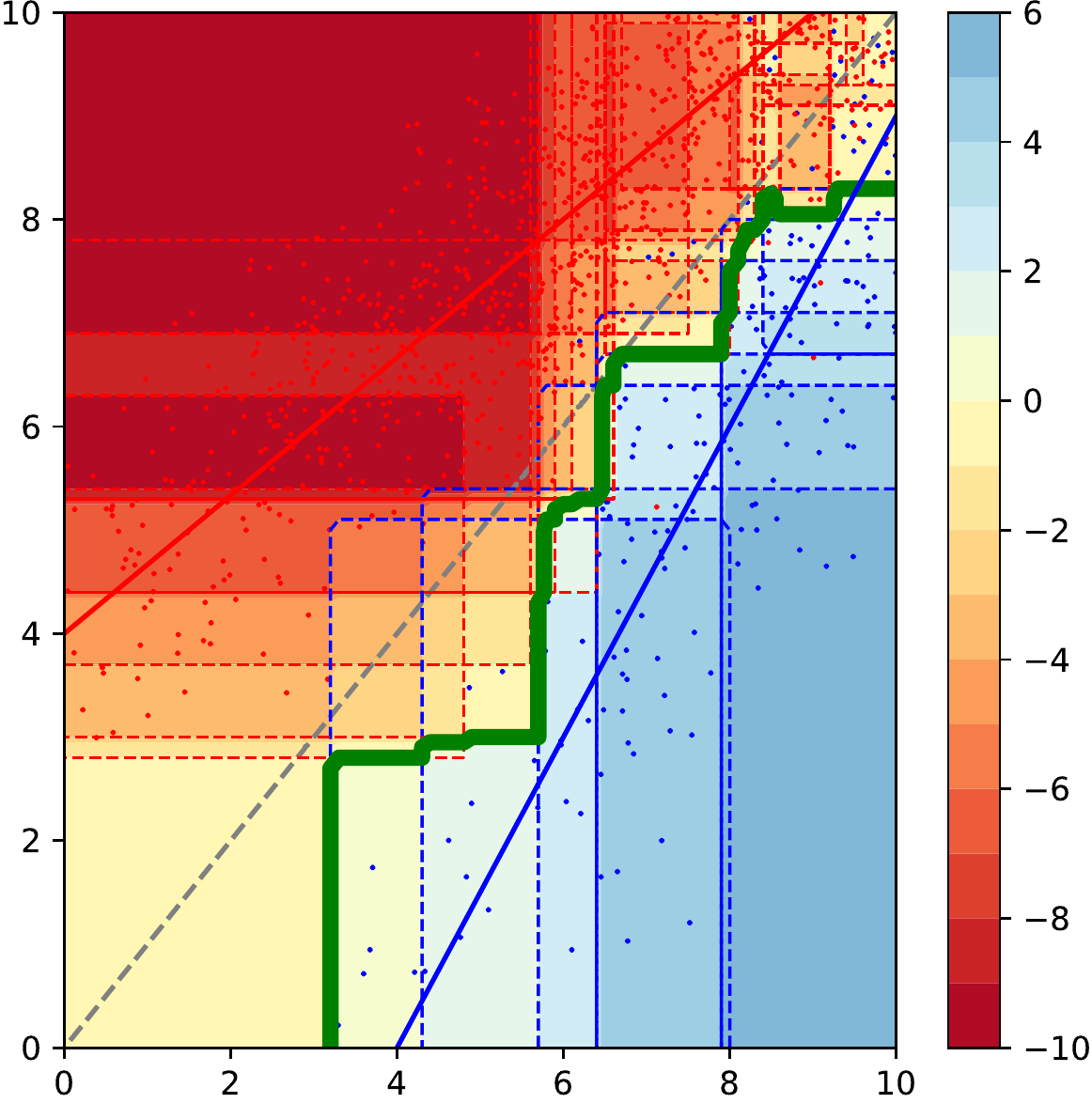}}\hfill%
\subfloat[all 4193 rules\label{fig:synthAll}]{\includegraphics[width=0.32\linewidth,height=32mm]{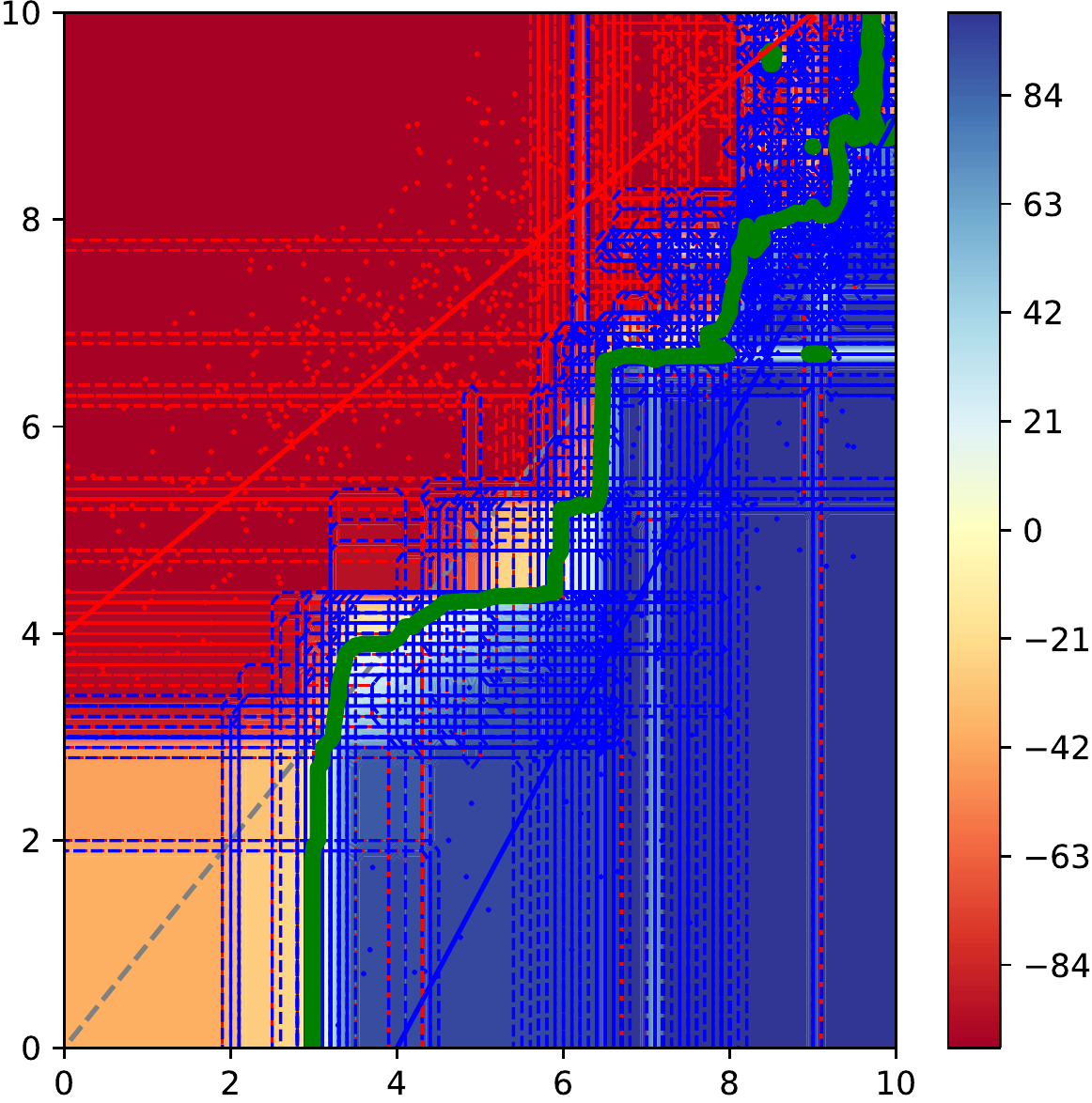}}%
  \caption{Synthetic data with 800 red and 200 blue instances, normally distributed along the corresponding lines. The rectangles denote the rules, their colors refer to the difference between blue and red votes. The green line visualizes the decision boundary.}
    \label{fig:synth}
\end{figure}

As a first step we applied our method on a synthetic data set, where the task is to approximate an oblique decision boundary in two-dimensional space (Fig.~\ref{fig:synth}). The results show that selecting the best rules according to a certain heuristic (m-estimate in this case) yields a set of redundant rules, covering a mostly homogeneous region. In contrast, weighted covering selects more diverse rules, covering the instance space more evenly. This results in a more precise approximation of the original decision boundary.

\begin{figure}[t]
    \centering
\includegraphics[width=1.0\linewidth]{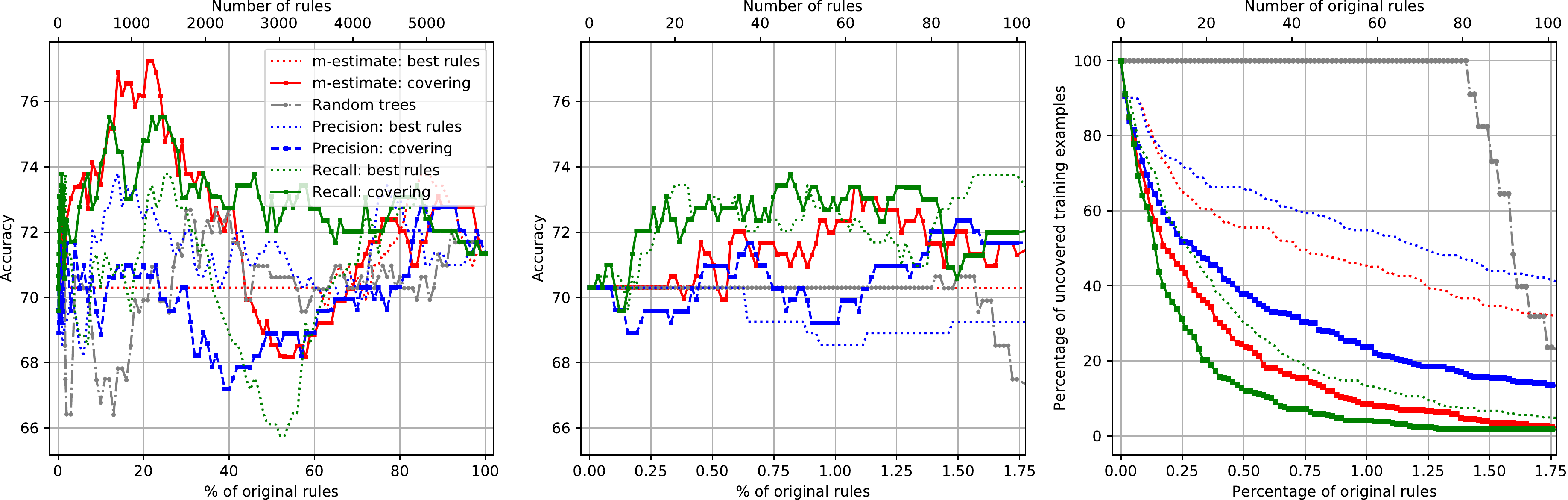}%
  \caption{Accuracy on the data set \emph{breast-cancer} using up to $d$ rules (left) or up to 100 rules (middle). For the latter, we also show the fraction of uncovered test instances (right). The approach ``Random trees'' adds all rules of a randomly selected tree at once.}
    \label{fig:breastcancer}
\end{figure}

Fig.~\ref{fig:breastcancer} shows the behavior of the different rule selection strategies on the \emph{breast-cancer} data set from the UCI repository (obtained via 10 fold cross validation). 
In particular weighted covering using the m-estimate or recall performs well, reaching the baseline performance of using all rules already after selecting around $20$ (recall) or $60$ (m-estimate) rules. When using around $20\%$ of all rules, a clearly visible improvement can be observed. Interestingly, the performance can also clearly drop again (m-estimate). Recall seems to be more robust, also when using the best rule strategy. It is also the approach with the steepest increase in coverage. Results on other data sets (not shown here) exhibit similar effects.

{}

\section{Conclusions}
We have proposed a technique for simplifying random forests based on rule subset selection. 
As our preliminary results show, high predictive performance is already achievable with a few rules, often even clearly outperforming the baselines.  
Our results suggest that the greatest potential for improvements lies in the rule selection strategy.

\enlargethispage*{12pt}

\bibliography{bibliography}
\bibliographystyle{splncs04}

\end{document}